# Development of a rule-based lemmatization algorithm through Finite State Machine for Uzbek language

Maksud Sharipov [1] and Og'abek Sobirov[1]

[1] *Urgench State University, Department of Information Technologies, 14, Kh.Alimdjon str, Urgench city, 220100, Uzbekistan*

**Abstract**
Lemmatization is one of the core concepts in natural language processing, thus creating a lemmatization tool is an important task. This paper discusses the construction of a lemmatization algorithm for the Uzbek language. The main purpose of the work is to remove affixes of words in the Uzbek language by means of the finite state machine and to identify a lemma (*a word that can be found in the dictionary*) of the word. The process of removing affixes uses a database of affixes and part of speech knowledge. This lemmatization consists of the general rules and a part of speech data of the Uzbek language, affixes, classification of affixes, removing affixes on the basis of the finite state machine for each class, as well as a definition of this word lemma.

**Keywords**
Uzbek language, lemma, affix, lemmatization, part of speech, finite state machine

## 1. Introduction

In linguistics, a lemma is a "dictionary form" or a "canonical form" of words. More specifically, a lemma is the canonical form of a *lexeme* where it refers to the set of all the forms that have the same meaning, and a lemma refers to the particular form that is chosen as a base form to represent the lexeme. Lemmas have special significance in highly inflected and agglutinative languages, such as Uzbek [1]. Uzbek is a low-resource agglutinative language [2], whose words are generated by adding affixes to the root forms. Affixes: prefixes, suffixes and infixes to the stem of the given word or words [3]. Lemmatization is an important step in many natural language processing (NLP) tasks, such as information retrieval, and information extraction [4]. Lemmatization is the process of determining the lemma for a given word, so different inflected forms of a word can be analyzed as a single item, and it creates the set of lemmas of a lexical database. It is conceived as starting from text words found in a corpus and leading to lemmas heading dictionary entries [1,4]. We present in this paper how to create a lemmatization algorithm for the Uzbek language and some examples. Our lemmatization method shows good performance in Uzbek, and it is easily extensible to other agglutinative languages.

Stemming is the process of determining the unchanging stem of a given word (stemma), which does not necessarily coincide with its morphological root [5]. But there are differences between them, let us we will consider it with the following word, *o'qigan (read (past participle))*
- *o'q/i/gan* stemming shows: *o'q (arrow)*
- *o'q/i/gan* lemmatization shows: *o'qimoq (to read)*, *-moq* means *to*

There has only little work been done in the field of Uzbek lemmatization, thus, in this paper, we propose a lemmatization algorithm for the Uzbek language.

## 2. Related work

Valuable amount of work has been done to solve different stemming-related NLP problems, and we can see both stemming and lemmatization for other languages. The lemmatization method has shown good potential in many diverse languages and was further developed on larger datasets of the tested languages, with no exception to Asian languages [4]. One of the works done on stemming that is really close to Uzbek, is the Turkish one, where it has been proposed two lemmatization modules, a static (offline), and a dynamic (online) module. The static module is constructed once and new words originating from new sentences crawled from any source can be added to the offline module by using the online module. The online module has the ability to new relation types between the words and lemmas that can be allowed to apply a suitable illustration method to solve obscurity [6]. We can also see these works in the Uzbek language in this field as well. It starts with an analysis of scientific research related to the creation of morphological analyzers belonging to the family of agglutinative languages. The functionality of Uzbek morphological analyzers has also been studied [7]. Another paper, which deals with affix stripping morphological analyzer is developed for the Uzbek language which is based on the morphotactic rules of the language. This model focuses on reaching the stem of a word without using any lexicon while making the morphological analysis [8].

Apart from stemming-related works on low-resource Uzbek language, recent years have seen an increasing trend in NLP works on Uzbek language, such as sentiment analysis [9], stopwords dataset [10], as well as cross-lingual word embeddings [11].

Normalization, namely, word lemmatization is a one of the main text preprocessing steps needed in many downstream NLP tasks. The lemmatization is a process for assigning a lemma for every word form [12]. It is the process of assembling the inflected parts of a word such that they can be recognized as a single element, called the word's lemma or it's vocabulary form. This process is the same as stemming but it adds meaning to particular words. In simple words, it connects text with alike meanings to a single word [13].

## 3. Methodology

In our approach, the methodology behind this process is carried out in two steps, from the moment we are given the 'raw text' as follows: First, text normalization, raw text is divided into word tokens, then these consisting only of punctuation and numbers are deleted. In texts, the total number of words called *tokens* [14]. Second, in order to establish the lemmatization algorithm, two databases have been designed and implemented: Respectively, we named databases as **Words** and **Affixes**. The aim of creating those two databases is to use them when finding the lemma of words and removing affixes. In this process, the removal of affixes is based on a finite state machine (FSM) technology. The FSM consists of a set of states $s_i$ and a set of transitions between pairs of states $s_i, s_j$ [15]. FSM is created based on the rules of word structure in the Uzbek language and uses an affixes database when removing affixes.

### 3.1. Word structure

A word consists of morphemes, the morpheme is the smallest meaningful part of a word structure [16]. For example, *kitob/lar/da/gina (only in books)*. *Kitoblardagina* is word, *kitob, -lar, -da* and *-gina* are morphemes. These parts are present in different words and they have different meanings. If these parts are divided again, their meaning will be lost. There are two types of morphemes: free morphemes and bound morphemes [Figure 1]. The free morpheme is the main part of the word meaning, and bound morphemes are used in conjunction with this free morpheme. Accordingly, the free morpheme is also called a base (root), and the bound morphemes are also called affixes [16,17,18]. Therefore, as mentioned in the example above, *kitob (book)* is a free morpheme, *-lar, -da* and *-gina* are bound morphemes. The base holds the meaning, and the affixes help to shape the meaning to the desired form.

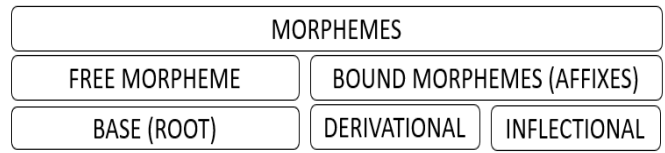

**Figure 1**: Morpheme structure

There are two types of bound morphemes: derivational and inflectional. Derivational affixes change the word's meaning. Inflectional affixes change the form of the word. The free morpheme is the basis for both word formation and form formation. For example, *til/shunos (linguist)* there are *til (language)* is base, *-shunos* is derivational affix and *til/lar (languages)* there are *til (language)* is base, *-lar* is the inflectional affix. In the Uzbek language, inflectional affixes are also divided into two types: lexical and grammatical. Affixes are used in Uzbek mostly as suffixes and are added after the base of the word. Sometimes they are added to the base as a prefix as well [16,17,18].

**Table 1** Word structure in Uzbek language

| Word | Morphemes | | | | |
|---|---|---|---|---|---|
| | **Prefix** (consists derivational or lexical suffixes) | **Base (Root)** | **Derivational suffixes** | **Inflectional suffixes** | |
| | | | | **Lexical suffix** | **Grammatical suffix** |
| *paxtakorlarga (to cotton growers)* | | *paxta (cotton)* | *kor* | *lar* | *ga* |
| *serhosil (fertile)* | *ser (derivational affix)* | *hosil (crop)* | | | |
| *nimpushti (pinker)* | *nim (lexical affix)* | *pushti (pink)* | | | |

As can be seen above in the Table 1, following examples can be shown regarding the formation of words from lemmas:
- Lemma [base]: *yurmoq (to walk), munosabat (attitude), tez (fast)*
- Lemma [prefix+base]: *be/foyda (useless), ham/kasb (colleague), no/umid (hopeless)*
- Lemma [base+derivational]: *yangi/lamoq (to renew), aylan/a (circle), ayb/dor (guilty)*
- Lemma [base+lexical]: *koʻr/satmoq (to show), koʻk/ish (blue-ish), ikki/nchi (the second)*
- Lemma [base+derivational+lexical]: *ikki/lan/moq (hesitation)*

### 3.2. Part of Speech and Affixes

In the second step of the algorithm uses a part of speech (POS) tag and its database [*database name*: Words] to perform its task. POS is divided into classes according to what a question of words is and how they express a generalized meaning. In the Uzbek language, POS is divided into three classes [19] open word classes, closed word classes, and intermediate words [Table 3]. The work begins by searching for the token in the WORDS database. In the database, each POS tag is located in different forms, such as a lemma. Some POS words do not have an affix, like closed word classes and some intermediate words. Therefore, it is necessary to search for it first before attempting to remove the affix from the token content. If the token is not found in the database, the affixes in the token should be removed one by one based on the word structure of the Uzbek language and searched in this database at each while.

**Table 2** POS tags in Uzbek language

| PART OF SPEECH | EXAMPLE |
|---|---|
| Open word classes (lexical) | |
| Fe'l (Verb) | *yugurmoq (to run), bormoq (to go)* |
| Olmosh (Pronoun) | *men (I), biz (we)* |
| Ot (Noun) | *avtomobil (a car), olma (an apple)* |
| Ravish (Adverb) | *rostdan ham (really), bugun (today)* |
| Sifat (Adjective) | *baxtli (happy), katta (big)* |
| Son (Numeral) | *bir (one), qirq besh (forty-five)* |
| Closed word classes (grammatical) | |
| Bog'lovchi (Conjuction) | *va (and), yoki (or)* |
| Ko'makchi (Auxiliary) | *uchun (for), bilan (with)* |
| Yuklama (Particle) | *ham (also), faqat (only)* |
| Intermediate words | |
| Modal (Modal) | *afsuski (unfortunately), darhaqiqiat (actually)* |
| Taqlid (Imitation) | *cheep-cheep, buzz* |
| Undov (Interjection) | *ooh! bah!* |

As mentioned above, affixes are used in Uzbek mainly as suffixes and are added after the base of the word. There are also prefixes that consist of derivational and lexical suffixes. Affixes exist in open word classes other than pronouns [Table 3]. But when a pronoun is used in place of others, it can take their affixes. Closed word classes never get affix and do not replace open word ones. Intermediate words also do not get affixes, but some may replace open word classes [2,16,20]. In the table below, we have given the number of affixes that we have collected from the works of literature.

**Table 3** Affixes in Uzbek language

| Base (Root) | Derivational suffixes [suffixes (allomorphs)] | Lexical suffixes [suffixes (allomorphs)] | Grammatical suffixes [suffixes (allomorphs)] |
|---|---|---|---|
| Fe'l (Verb) | 26 (33) | 36 (64) | 35 (47) |
| Ot (Noun) | 90 (103) | 21 (26) | 14 (21) |
| Ravish (Adverb) | 18 (22) | | |
| Sifat (Adjective) | 58 (75) | 4 (5) | |
| Son (Numeral) | | 13 (14) | |

## 3.3. FSM for lemmatization

FSM [Figure 2], which removes affixes for lemmatization, works in the following order. Since there are no grammatical suffixes in the lemma, we remove such verb and noun suffixes according to table-3 directly from the word. Typically, the algorithm removes a single affix when referring to FSM, however, in this state, the FSM removes all of these suffixes at once, not just one (n – removing multiple affixes). Let's look at an example of a noun: *kitob/im/ning (of my book)*, there are *kitob (book)* lemma, *-im* and *-ning* are grammatical suffixes and they will be removed in one state. If this type of affix does not exist in the word, goes directly to the previous lexical suffix removal [0 – empty transaction].

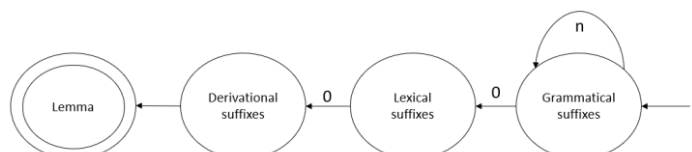

**Figure 2**. FSM schema

## 4. Algorithm creation

In this section, the lemmatization algorithm is developed based on rules through FSM for the Uzbek language [Figure 3]. The flowchart of the lemmatization algorithm is shown in Figure 4. If the raw text was written without any spell mistakes and is in the modern Uzbek language, the algorithm works with high accuracy.

```
Step1: Input raw text
Step2: Split the input to token through word tokenize
Step3: If the token is punctuated (like: .,!) or number (like:
       12, 2022), remove it
Step4: Search the token from WORDS database, if found it, pass
       step6, if it not fount, pass step5
Step5: Remove the affix from the token content based on the FSM
       with AFFIXES database, and pass step4
Step6: Output token as a lemma
```

**Figure 3**. Steps performed by the proposed lemmatization algorithm.

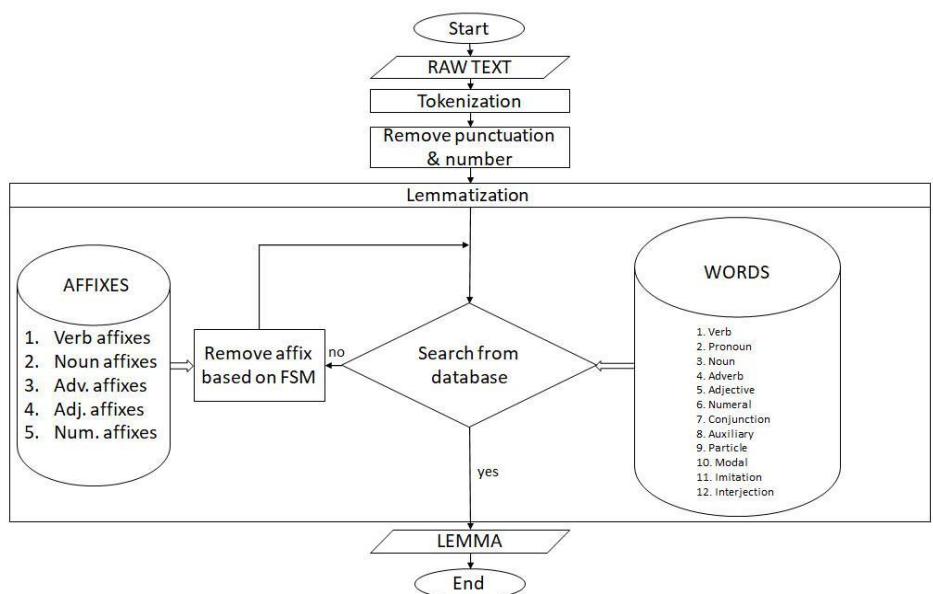

**Figure 4**. Flowchart of Lemmatization Algorithm

## 5. Conclusion and future work

In this paper, we presented a database: ***Words,*** that have been created based on word classes, and the database of affixes: ***Affixes***. Furthermore, the FSM was created to remove affixes. Based on this, the lemmatization algorithm was developed for the Uzbek language. It is aimed to use in the NLP tasks of Uzbek. In the future, our aim is to increase the reliability of the algorithm by expanding the words database.

It is planned to carry out this work in a Uzbek corpus [21] which has more 80 000 words and word phrases.

Moreover, we aimed to create a program for it as a python package then it will place in pypi.org, like pip install UzbekLemmatizer and we also place its source code on the github.com platform. This is because, so far there is no such package for the Uzbek language in python packages, even algorithm is no.